\documentclass[review]{elsarticle}

\usepackage{graphicx}
\usepackage{amsmath}
\usepackage{multirow}
\usepackage{tabularx}
\usepackage{subfig}
\usepackage{caption}

\begin{document}
\begin{frontmatter}

\title{Cross Domain Knowledge Learning with Dual-branch Adversarial Network for Vehicle Re-identification}

\author{Jinjia Peng}
\ead{jinjiapeng@dlmu.edu.cn}

\author{Huibing Wang\corref{mycorrespondingauthor}}
\ead{huibing.wang@dlmu.edu.cn}

\author{Xianping Fu\corref{mycorrespondingauthor}}

\ead{fxp@dlmu.edu.cn}
\address{College of Information and Science Technology, Dalian Maritime University, Danlian, Liaoning, 116021, China}

\cortext[mycorrespondingauthor]{Both corresponding authors}

\begin{abstract}
The widespread popularization of vehicles has facilitated all people's life during the last decades. However, the emergence of a large number of vehicles poses the critical but challenging problem of vehicle re-identification (reID). Till now, for most vehicle reID algorithms, both the training and testing processes are conducted on the same annotated datasets under supervision. However, even a well-trained model will still cause fateful performance drop due to the severe domain bias between the trained dataset and the real-world scenes.

To address this problem, this paper proposes a domain adaptation framework for vehicle reID (DAVR), which narrows the cross-domain bias by fully exploiting the labeled data from the source domain to adapt the target domain. DAVR develops an image-to-image translation network named Dual-branch Adversarial Network (DAN), which could promote the images from the source domain (well-labeled) to learn the style of target domain (unlabeled) without any annotation and preserve identity information from source domain. Then the generated images are employed to train the vehicle reID model by a proposed attention-based feature learning model with more reasonable styles. Through the proposed framework, the well-trained reID model has better domain adaptation ability for various scenes in real-world situations. Comprehensive experimental results have demonstrated that our proposed DAVR can achieve excellent performances on both VehicleID dataset and VeRi-776 dataset.

\end{abstract}

\begin{keyword}
\texttt{Domain adaptation  \sep Dual-branch adversarial network  \sep Vehicle re-identification}
\end{keyword}

\end{frontmatter}

\section{Introduction}
Recently, video surveillance for traffic control and security is playing a growing influence on current public transportation systems. During the last decade, vehicle-related researches have attracted more interest and made great progress in computer vision community, such as vehicle detection \cite{ref_article1}\cite{ref_article2}, segmentation \cite{ref_article3}\cite{ref_article4} and classification \cite{ref_article5}\cite{fang2018dart}. Different with the tasks above, vehicle reID aims to precisely match a certain vehicle across scenes captured from multiple non-overlapping cameras, which plays a crucial role in constructing the smart cities \cite{ref_article7}. Meanwhile, vehicle reID can be automatically carried out with less time consuming and manual labor. Therefore, vehicle reID is of vital significance for intelligent transport and arouses attentions from researchers all over the world.

\begin{figure}
\centering
\includegraphics[width=8cm]{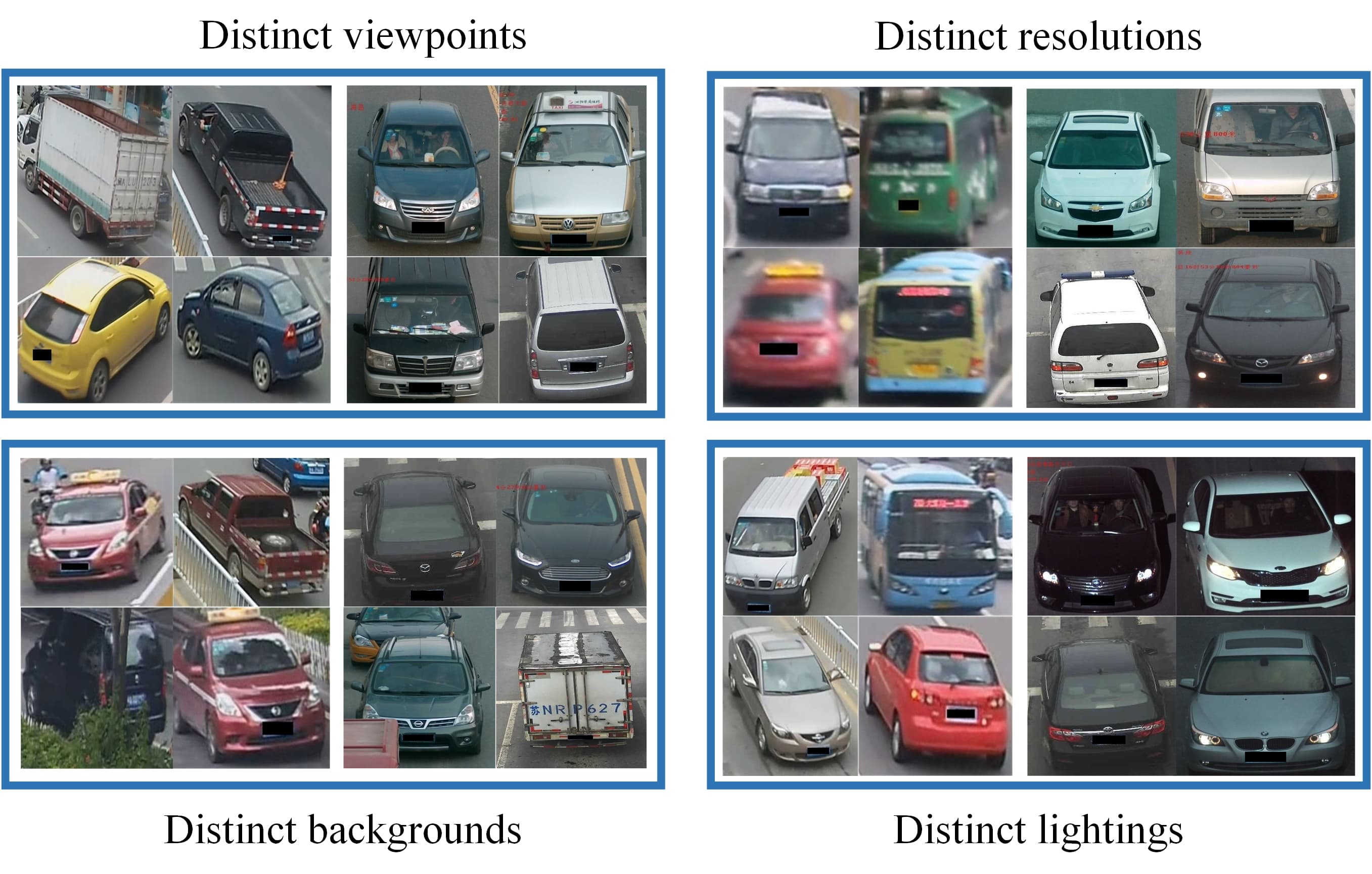}
\caption{Illustration of the domain bias between different domains. For every group, the images of left side are from VeRi-776 while the ones of right side are from VehicleID.} \label{fig1}
\end{figure}

Even though some progress has been made for vehicle reID, how to design an excellent algorithm to adapt domain bias between different scenes still matters the whole system. The fatal reason is that vehicle reID always faces heterogeneous real-world scenes which contain intensive changes in illuminations and backgrounds. Therefore, the same vehicle captured in various scenes may present different visual appearances, which poses great challenges for the task of vehicle reID. What's more, for one domain, it could not contain all cases in real scenario, which makes different domains have their own unique style and causes the bias between domains. Usually, as shown in Fig.\ref{fig1}, domains differ form each other regarding lightings, viewpoints and backgrounds, even the resolution. It is observed that, when the well-trained reID model is tested on other domain without fine-tuning, there is always a severe performance drop due to the domain bias. However, most existing works on reID follow the supervised learning paradigm which always trains the reID model using the images in the target domain first to adapt the style of the target domain \cite{wang2017effective}\cite{ref_article9}\cite{ref_article10}\cite{ref_article34}\cite{ref_article35}\cite{wang2018multiview}. Hence, most of these supervised learning methods can not be utilized in the real scenario directly. Furthermore, since most images in video surveillance are not labeled and annotating large-scale datasets is prohibitively expensive, research efforts \cite{ren2019semi}\cite{sahoo2018meta} are desired to narrow-down or eliminate the domain bias.

Different with the supervised reID task which has been explored by various works \cite{wu20193}\cite{wu2019cross}, there are a few studies on reID about the cross-domain adaptation. And only several methods exploit unlabeled target data for unsupervised person reID modelling \cite{ref_article13}\cite{ref_article14}\cite{ref_article16}\cite{wu2019cycle}. However, some of them need extra information about source domain while training, such as attribute labels and spatio-temporal labels, which do not exist on some datasets. And there are only several methods exploiting unsupervised learning without any labels, such as SPGAN \cite{ref_article13} and PTGAN \cite{ref_article17}. SPGAN is designed for person reID that integrates a SiaNet with CycleGAN \cite{ref_article21} and it does not need any additional labels during training. However, though SPGAN is effective on the person transfer task, it causes deformation and color distortion in vehicle transfer task in our experiment. PTGAN is composed of PSPNet \cite{ref_article22} and CycleGAN to learn the style of target domain and maintain the identity information of source domain. In order to keep the identity information, PSPNet is utilized to segment the person images first. It needs pre-trained segmention model for PSPNet, which increases the complexity of the training stage.

To sum up, most existing domain adaptation methods usually need special annotations or complex training process, which cannot be utilized for vehicle reID task. Therefore, in this paper, we propose DAVR which employs an end-to-end image-to-image translation network DAN, meanwhile the generated images are utilized to train reID model by an proposed feature learning network. In DAVR, DAN employs different branches to train the content encoder and style encoder without any annotations to preserve the identity information of images from source domain and learn style of images from target domain. Besides that, for adapting the target domain (unlabeled) and having better generalization ability, an attention-based network (ATTNet) is designed in DAVR to train the vehicle reID model with the images generated from DAN. In summary, our contributions can be summarized into three aspects:

1) We propose DAVR to optimize the reID model that is trained by labeled source domain to adapt the unlabeled target domain, which contains DAN for generating images and ATTNet for better training reID model utilizing the generated images.

2) In DAVR, DAN is proposed to generate the images which have the style of target domain and preserve identity information of source domain. It is an efficient unsupervised learning model and works by transferring content and style between different domains separately.

3) To better train reID model with the images generated by DAN, ATTNet is presented in DAVR, which is based on attention structure and could extract more distinctive cues while suppressing background for vehicle reID task.

The rest of this paper is organized as follows. In Section 2, we review and discuss the related works. Section 3 illustrates the proposed method in detail. Experimental results and comparisons on two vehicle reID datasets are discussed in Section 4, followed by conclusions in Section 5.

\section{Related Work}

In this section, we briefly review the methods of image-image translation, vehicle reID and cross-domain person reID methods.

\subsection{Image-image Translation Methods}
Image-image translation aims at constructing a mapping function between two domains. In recent years, a lot of studies based on generative adversarial networks have shown remarkable performance improvement. Most of them utilized paired training data to produce impressive image-to-image transition results. For instance, pix2pix \cite{ref_article36} employed a conditional GANs to learn mappings from input to output images by combining adversarial loss and $L_1$ loss. However, it is difficult to acquire the paired training data while easier to collect data without data, consequently unsupervised image translation is more applicable. For these condition, several methods have been proposed, such as CycleGAN \cite{ref_article21} and DiscoGAN \cite{ref_article21}. They preserved key attributes by cycle consistency loss and transferred style between two domains.

\subsection{Vehicle ReID Methods}
With the prosperity of deep learning, feature learning by deep networks has become a common practice in vehicle reID tasks. Zapletal el at. \cite{ref_article39} extracted a full-edged 3D bounding box of vehicles and then utilized color histograms and histograms of oriented gradients to solve reID problem by a linear regressor. In \cite{ref_article11}, coupled cluster loss was proposed to minimumize intra distance to train the vehicle re-identification network. VAMI \cite{ref_article12} transformed single-view feature into a global multiview feature representation to better optimize the metric learning for training reID model. Spatial-temporal information is another important clue which should be considered for vehicle reID. For instance, In \cite{ref_article18}, besides considering the local region features of vehicle images, the spatial-temporal constrain is modeled by log-normal distribution. \cite{ref_article19} introduced a siamese-Cnn+Path-LSTM model to incorporate complex spatio-temoral information for regularizing the reID results. PROVID \cite{ref_article20} introduced the information of license plates, visual features and spatial-temporal relations with a progressive strategy to learn similarity scores between vehicle images. Bai el at. \cite{ref_article40} proposed a group sensitive triplet embedding for CNNs to deal with intra-class variance in learning representation and the mean-valued triplet loss was given to alleviate the negative impact of improper triplet sampling during training stage.

\subsection{Cross-domain Person ReID Methods}
For the problem of cross-domain reID, the biggest challenge is that it is difficult to maintain the image identity information when transferring the annotated images from source to target domain in an unsupervised method. Hence, to overcome this problem, SPGAN \cite{ref_article13} was designed to integrate of a siamese network and a CycleGAN to preserve the self-similarity about an image before and after translation, and domain-dissimilarity about a translated source image and a target image. TJ-AIDL \cite{ref_article16} introduced simultaneously learned an attribute-semantic and identity discriminative feature representation space transferable to any new target domain for reID tasks without the need for collecting new labeled training data from the target domain. PTGAN \cite{ref_article17} was designed for person transfer, which employed the PSPNet and CycleGAN to generate high quality person images and keep the person identities and style.

From the above, we could see that only a few studies focus on the problem of cross-domain person reID. And little works consider the domain adaptation about the vehicle reID. We have discussed the reason that the method of cross-domain adaptation about the person reID could not be utilized in the vehicle reID task in the previous section. Hence, in this paper, we propose a novel framework to adapt the cross domain bias for vehicle reID.

\section{Method}
\subsection{Overview}

\begin{figure}[htbp]
\includegraphics[width=\textwidth]{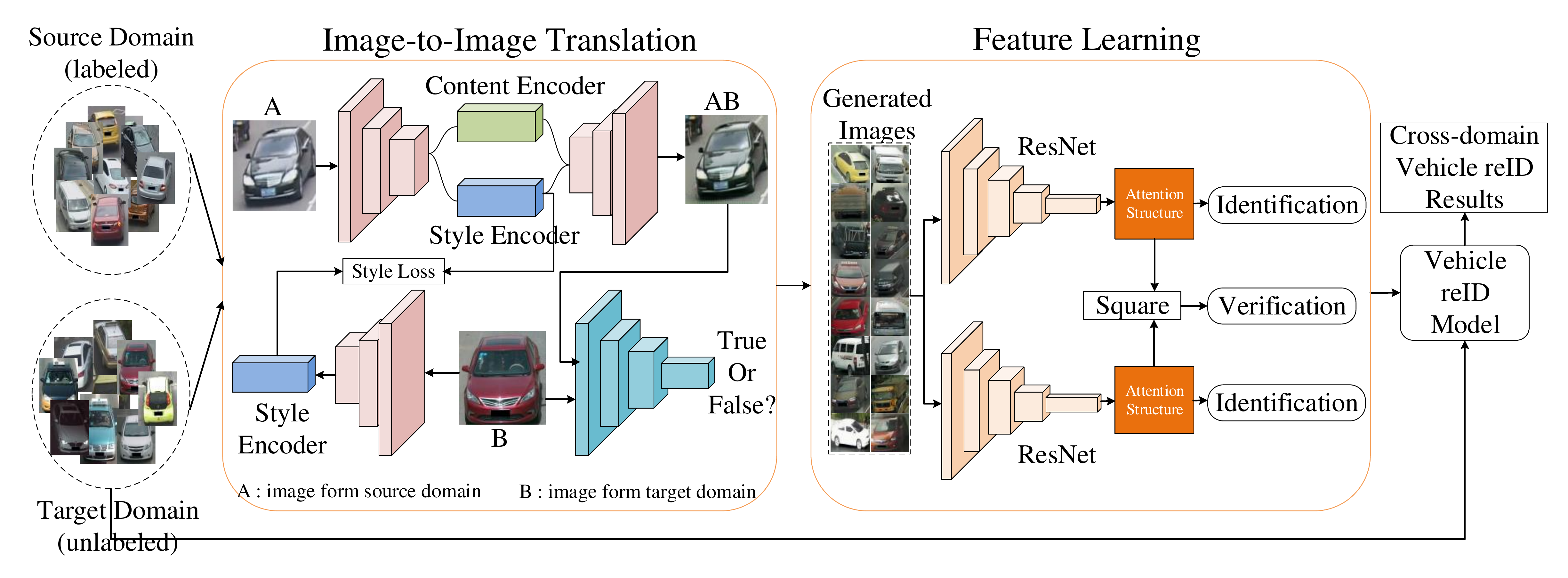}
\caption{The overview of DAVR. DAVR develops an image-to-image translation module to generate images with the style of target domain and then the generated images could be trained by an attention-based feature learning module.} \label{fig2}
\end{figure}

Our ultimate goal is to perform vehicle reID model in an unknown target domain which is not labeled directly. Hence, as shown in Fig.\ref{fig2}, DAVR is proposed which contains DAN for generating images and ATTNet for training reID model. Through DAN, it could obtain images which have the style of target domain and preserve the identity information of source domain. And then the style transferred images are employed to train the vehicle reID model by ATTNet.

In this section, we introduce our method from two aspects: an image-to-image translation network to learn transfer mappings for different datasets in Section 3.2 and an attention based multi-task feature learning network for better feature representation learning in section 3.3.

\subsection{DAN}

\begin{figure}[htbp]
\centering
\includegraphics[width=10cm]{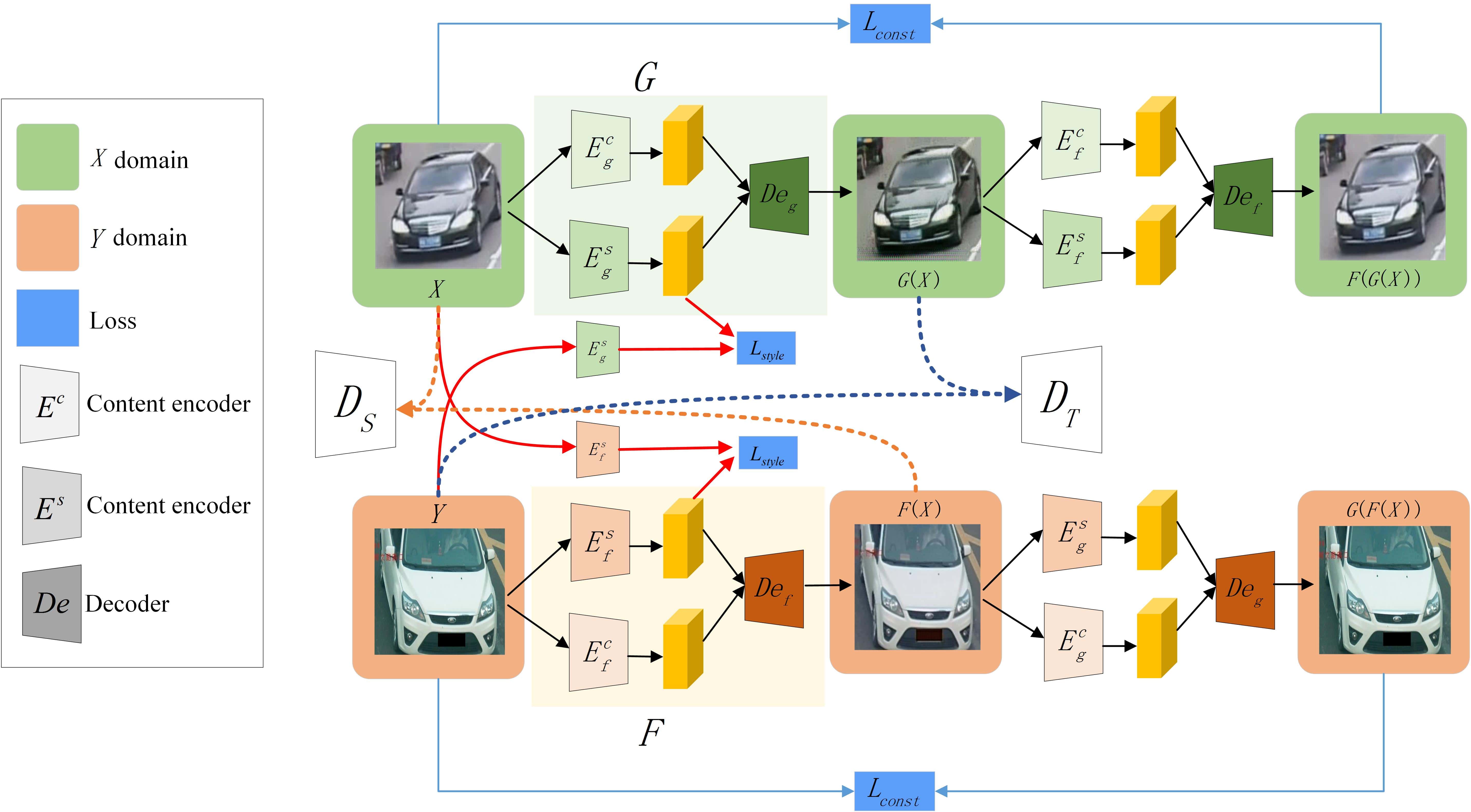}
\caption{The structure of DAN. DAN contains two mapping functions: $G:X{\rightarrow}Y$ and $F:Y{\rightarrow}X$, and associated adversarial discriminators $D_{T}$ and $D_{S}$. $L_{style}$ represents style loss which is employed to further regularize the mappings (best viewed in color).} \label{fig3}
\end{figure}
DAN is designed in DAVR to both transfer the style between source domain and target domain and preserve the identity information of images from source domain. As illustrated in Fig.\ref{fig3}, DAN consists of generators ${G,F}$, and domain discriminators ${D_{S}, D_{T}}$ for both domains. For each generator in DAN, it is composed of three components including content encoder ${E^c}$, style encoder ${E^s}$ and decoder ${De}$. ${E^c}$ is designed to preserve the identity information from images of source domain through the proposed attention model, which could extract the foreground while suppressing background. And to learn the style of target domain, the ${E^s}$ with the style loss is added to the translation network. At last, the decoder $De$ embeds the output of ${E^c}$ and $E^s$ to generate the translated image. Take domain ${X}$ as an example, the content encoder $E^c_{g}$ maps images onto a domain-invariant content space $(E^c_{g}:x\to{C_x})$ and the style encoder $E^s_{g}$ maps images onto the domain style space of ${Y}$ $(E^s_{g}:x\to{S_{y}})$. The generator $G$ generates images conditioned on both content and style vectors $(De:\{C_x,S_{y}\to{G(X)}\})$. The discriminator $D_{S}$ aims to discriminate real images and translated images in the domain $X$.
\begin{figure}[htbp]
\includegraphics[width=\textwidth]{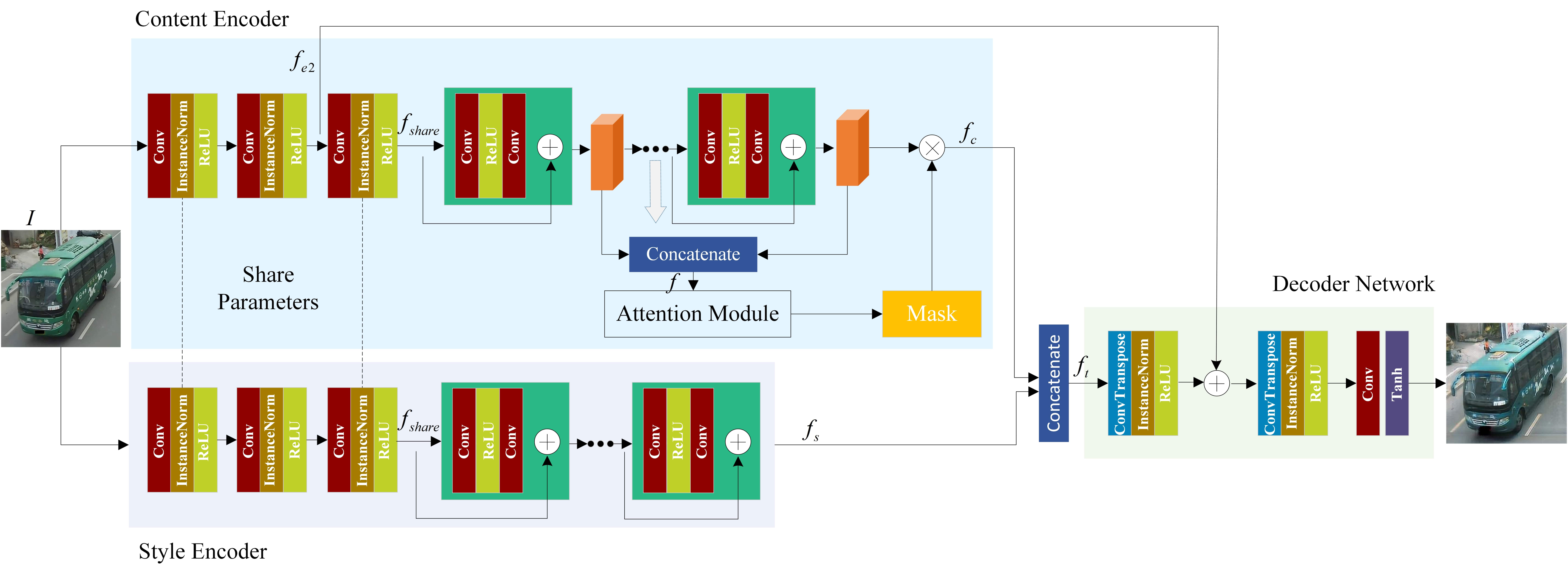}
\caption{Structure of Generator. Generator is composed of a content encoder network, a style encoder network and a decoder network (best viewed in color).} \label{fig4}
\end{figure}

\subsubsection{Content Encoder for Identity Preserving}

As shown in Fig.\ref{fig4}, the input image of the generator is defined as $I$. After 3 strided convolution blocks with stride $\frac{1}{2}$, we could obtain feature maps $f_{share}$. Based on the assumption that two domains share a common latent space, we share the weights of these 3 convolution blocks between $E^c_g$ and $E^c_f$. For every convolution block, it contains a convolutional layer, an instance normalization layer and a ReLU layer. Then $f_{share}$ is passed into subsequent network, which contains 9 residual blocks(ResBlocks) and the proposed attention model. In order to preserve the identity information from source domain, we remove the batch normalization layers in the original ResNet structure. And the attention model is designed to assign higher scores of visual attention to the region of interest while suppressing background.

As shown in Fig.\ref{fig4}, we denote the input feature map of attention model as $f$. In this work, a simple feature fusion structure is utilized to generate the $f$. All the outputs of the ResBlock are integrated to form $f$ which can be formulated as $f=[f_{r1}, f_{r2}, ..., f_{r9}]$, where $f_{ri}$ is the $ith$ feature map generated by the $ith$ ResBlock. $i\in [1,9]$ and $[\cdot]$ denotes the concatenation operation. For the feature vector $f_{i,j}\in{\Re^C}$ of the feature map at the spatial location $(i,j)$, we can calculate its corresponding attention mask $a_{i,j}$ by

\begin{equation}
a_{i,j}=Sigmod(FC(f_{i,j};W_{a}))
\end{equation}
where $FC$ is the Fully Connected layer (FC) to learn a mapping function in the attention module and $W_{a}$ are the weights of the FC. The final attention mask $\alpha=[a_{i,j}]$ is a probability map obtained using a Sigmoid layer. The scores represent the probability of foreground in the input image. And after the attention model, a mask $a$ is generated with higher scores for foreground. Hence, the attended feature map $f_{c}$ is computed by element-wise product of the attention mask and the input feature map, which could be described as follows:

\begin{equation}
f_{c(i,j)}=a_{i,j}\otimes{f_{i,j}}
\end{equation}
where $(i,j)$ is the spatial location in mask $a$ or feature map $f_{c}$. And $\otimes$ is performed in an element-wise product.

\subsubsection{Style Encoder for Learning Style Transfer Mappings}
As shown in Fig.\ref{fig4}, besides the content branch, there is a branch learning the style of target domain. In this branch, different with $E^c_g$ and $E^c_f$, the style network $E^s_g$ and $E^s_f$ do not contain the attention model. For instance, $E^s_g$ is composed of 3 convolution blocks, which are the same as the content encoder network, and 9 residual blocks(ResBlocks). The 3 convolution blocks share parameters with the content encoder network. To learn the style of the target domain, $E^s_g$ is designed with the style loss to output the style features $f_{s}$ that has similar distribution with the target domain $Y$. The style loss could be formulated as follows:

\begin{equation}
\begin{split}
L_{style}=  {\frac{1}{NM}{{(T(x)-A(y))^2}}}+{\frac{1}{NM}{{(T(y)-A(x))^2}}}
\end{split}
\end{equation}
where $N$ is the number of feature maps, $M$ is calculated by $width\times{height}$, $width$ and $height$ represent the width and height of images. $T(x)$, $T(y)$, $A(y)$ and $A(x)$ are the gram matrix of output features $E_{g}^s(x)$, $E_{f}^s(y)$, $E_{g}^s(y)$ and $E_{f}^s(x)$, respectively.

We calculate the style loss between images from source domain and target domain to compare differences of style between images. Thus images from different domains could learn the style from each other.

\subsubsection{Decoder Network for Embedding Two Stream Features}
For the decoder network, it is composed of 2 deconvolutional layers and a convolutional layer to output the generated images $G(I)$. As shown in Fig.\ref{fig4}, the input of the decoder network is the combination of $f_{c}$ and $f_{s}$ which represent the content features and style features, respectively. In this paper, we employ a concatenate layer to integrate $f_{c}$ with $f_{s}$ and a global skip connection structure to make training faster and resulting model generalizes better, which could be expressed as:

\begin{equation}
G(I)=tanh(conv(deconv(deconv([f_{c}, f_{s}])+f_{e2})))
\end{equation}
where $[.]$ represents the concatenate layer. And $f_{e2}$ represents the feature map generated by the 2th stride convolution blocks.

\subsubsection{Loss Function}
We formulate the loss function in DAN as a combination of adversarial loss, cycle consistency loss, identity loss and style loss. The adversarial loss and style loss guide the learning of the domain-migration network. The identity loss and cycle consistency loss preserve the semantic consistency and visual similarity of intra-class instances across domains. The objective function could be described as follows:
\begin{equation}
L=L_{adv}+ \lambda_{1}L_{cyc}+ \lambda_{2}L_{id}+\lambda_{3}L_{style}
\end{equation}
where the style loss $L_{style}$ could be calculated by Eq.(3).

In our paper, DAN applies adversarial losses to both mapping functions. For the generator $F$ and its discriminator $D_{T}$, the objective could be expressed as:
\begin{equation}
L_{T}(F,D_T,X,Y)=E_{x\sim p_{data}(x)}[D_T(x)]+E_{y\sim p_{data}(y)}[||D_{T}(F(y))-1||_1]
\end{equation}
where, $X$ and $Y$ represent the source domain and target domain, respectively. $p_{data}(x)$ and $p_{data}(y)$ denote the sample distributions in the source and target domain. The objective of generator $G$ and discriminator $D_{S}$ also could be built.

Besides, the DAN requires $F(G(x))\approx x$ and $G(F(y))\approx y$ when it learns the mapping of $F$ and $G$. So the cycle consistency loss is employed in DAN which could make the network more stable. The cycle consistency loss could be defined as:
\begin{equation}
L_{cyc}(F,G,X,Y)=E_{x\sim p_{data}(x)}[||F(G(x))- x||_1]+E_{y\sim p_{data}(y)}[||G(F(y))- y||_1]
\end{equation}

DAN utilizes the target domain identity constraint as an auxiliary for image-image translation. Target domain identity constraint is introduced by \cite{ref_article27} to regularize the generator to be the identity matrix on samples from target domain, described as:
\begin{equation}
L_{id}(G,F,X,Y)=E_{y\sim p_{data}(y)}||F(y)- y||_1+E_{x\sim p_{data}(x)}||G(x)- x||_1
\end{equation}

\subsection{ATTNet}
The purpose of feature learning module is to obtain discriminative features that could be utilized for vehicle reID. And in order to make the reID model adapt to the target domain, it is better to focus on the meaningful parts of vehicle images and neglect the background when training the feature learning model. Hence, ATTNet which contains a two-stream reID network with attention structure is designed in this paper.

\begin{figure}
\includegraphics[width=\textwidth]{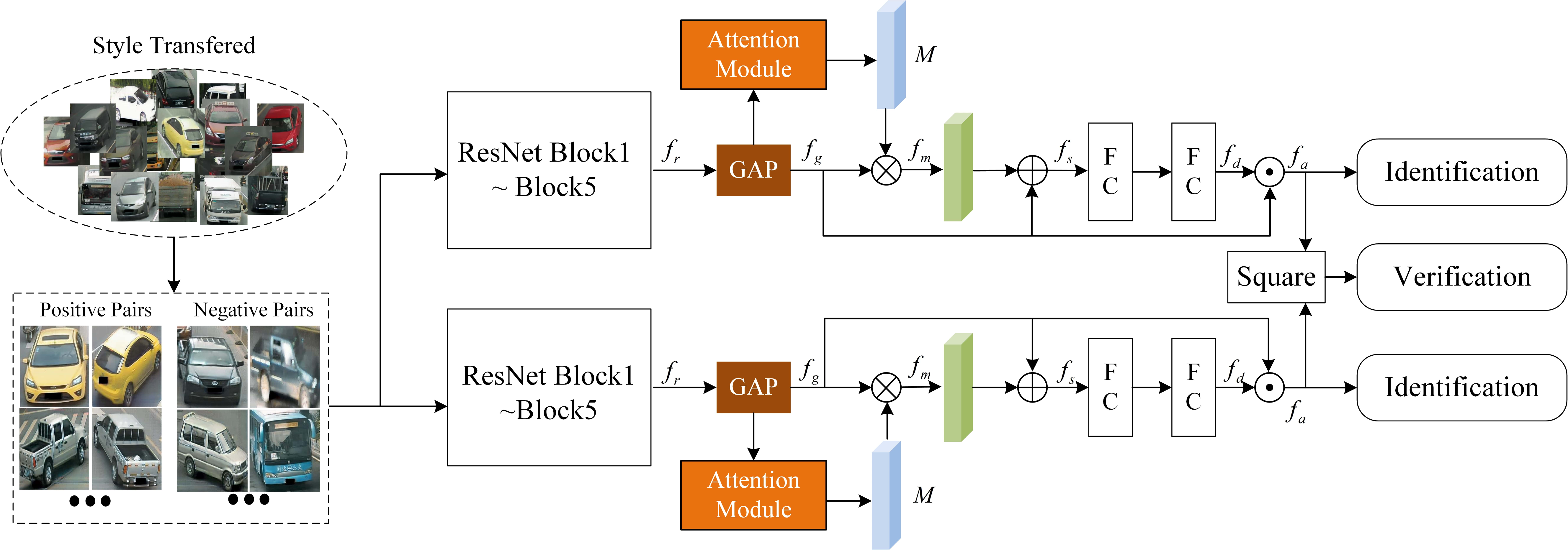}
\caption{The structure of ATTNet.} \label{fig5}
\end{figure}

ATTNet which includes identification network and verification network is a dual-branch structure and shares parameters. As shown in Fig.\ref{fig5}, the images from the generation module are divided into positive and negative samples pairs as inputs for ATTNet. Images with the same vehicle IDs are positive sample pairs, otherwise, they are defined as negative sample pairs. For one branch, the input image is fed into 5 ResNet Blocks \cite{ref_article37} to output the feature maps $f_{r}$ with the size of $7\times 7\times 2048$. Then they are passed into a Global Average Pooling (GAP) layer to obtain the feature map $f_{g}$. $f_{g}$ is utilized to generate the mask $M$ through the proposed attention structure. Given the feature map $f_{g}$, its attention map is computed as:

\begin{equation}
M = Softmax(Conv(f_{g}))
\end{equation}
where the one $Conv$ operator is $1\times 1$ convolution. After obtaining the attention map $M$, the attended feature map could be calculated by $f_m = f_{g}\otimes M$. The operator $\otimes$ is performed in an element-wise product. Then the attended feature map $f_{m}$ will be fed into the subsequent structure. However, since several training images may be spatial misalignment, the obtained attention mask $M$ could be somewhat imprecise and the attended feature map $f_{m}$ may be disturbed by noise. This will lead to the $f_{m}$ fail to contain some useful information in the original images. In order to solve this problem, a shortcut connection architecture is introduced to embed the input of the attention network directly to its output with an element-wise sum layer, which could be described as $f_s = f_{g} + f_{m}$. In this way, both the original feature map and the attended feature map are combined to form features $f_{s}$ and utilized as the input for the subsequent structure. After two FC layers, we could obtain the feature $f_{d}$. At last, a skip connection structure is utilized to integrate $f_{g}$ and $f_{d}$ by the concatenate layer to obtain more discriminative features for identification task and verification task, which could be described as $f_{a} = [f_{d}, f_{g}]$.

\section{Experiments}

In this section, we make an attempt to give a detailed analysis to demonstrate the effectiveness of our method. And the proposed DAVR is evaluated utilizing the mean average precision (mAP) and the Cumulative Match Characteristic (CMC) curve widely adopted in vehicle reID. First, we compare the generated images of DAN in DAVR with state-of-the-art methods. Then, the generated images are utilized to train the vehicle reID model by different methods to explore the effectiveness of proposed ATTNet. We conduct various experiments on two popular vehicle reID datasets: VeRi-776 and VehicleID.

\subsection{Datasets and Evaluation Metrics}

\subsubsection{Datasets.}
VeRi-776 \cite{ref_article20} is a large-scale urban surveillance vehicle dataset for reID, which contains over 50,000 images of 776 vehicles with identity annotations, camera geo-locations, image timestamps, vehicle types and color information. In this paper, 37,781 images of 576 vehicles are employed as a train set and 11,579 images of 200 vehicles are employed as a test set. A subset of 1,678 images in the test set generates the query set.

VehicleID \cite{ref_article11} is a surveillance dataset from the real-world scenario, which contains 26267 vehicles and 221763 images in total. From the original testing data, four subsets, which contain 800, 1600, 2400 and 3200 vehicles, are extracted for vehicle search in different scales. During testing, one image is randomly selected from one identity to obtain a gallery set with 800 images, then the remaining images are all employed as probe images. Three other test sets are processed in the same way.

\begin{figure}[ht]
\centering
\includegraphics[width=9cm]{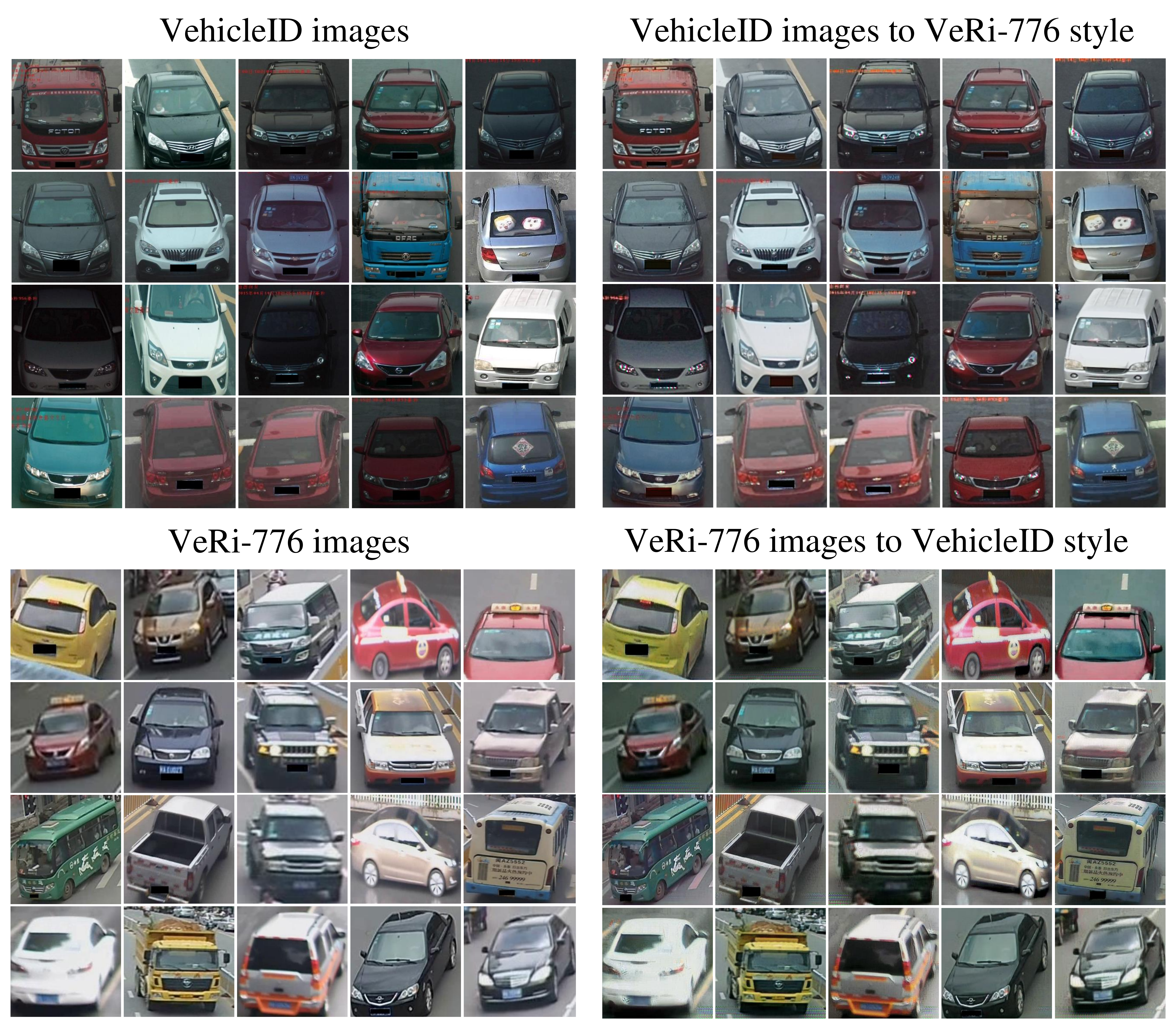}
\caption{Sample images of (upper left:) VehicleID dataset, (lower left:) VeRi-776 dataset, (upper right:) VehicleID images which are translated to VeRi-776 style, and (lower right:) VeRi-776 translated to VehicleID style.} \label{fig6}
\end{figure}

\subsubsection{Evaluation Metrics.}
For the vehicle reID task, we utilize CMC curve and mAP to evaluate the reID model. For each query, its average precision (AP) is computed from its precision-recall curve \cite{ref_article29}. And mAP is the mean value of average precisions across all queries.

\subsection{Implementation Details}
\subsubsection{Image-to-image Translation.}
For the translation module, we train the model in the tensorflow \cite{ref_article30} and the learning rate is set to 0.0002. It is worthy to note that we do not utilize any label notation during the learning procedure. The min-batch size of the proposed method is 16 and epoch is set to 6. 
During the testing procedure, we employ the Generator $G$ for VeRi-776 $\to$ VehicleID translation and the Generator $F$ for VeRi-776 $\to$ VehicleID translation. The translated images are utilized for training reID models.

\subsubsection{Feature Learning.}
For the feature learning module, We implement the proposed vehicle reID model in the Matconvnet \cite{ref_article31} framework. We utilize stochastic gradient descent with a momentum of $\mu=0.0005$ during the training procedure. The learning rate of the first 50 epoch is set to 0.1 while the last 5 to 0.01. As the scale of dataset can be quite large, training data is randomly divided into mini-batches with a batch size of 16.

\subsection{Evaluation}

\subsubsection{Comparison of Generated Images with Different Methods}
\begin{figure}[htbp]
\centering
\includegraphics[width=\textwidth]{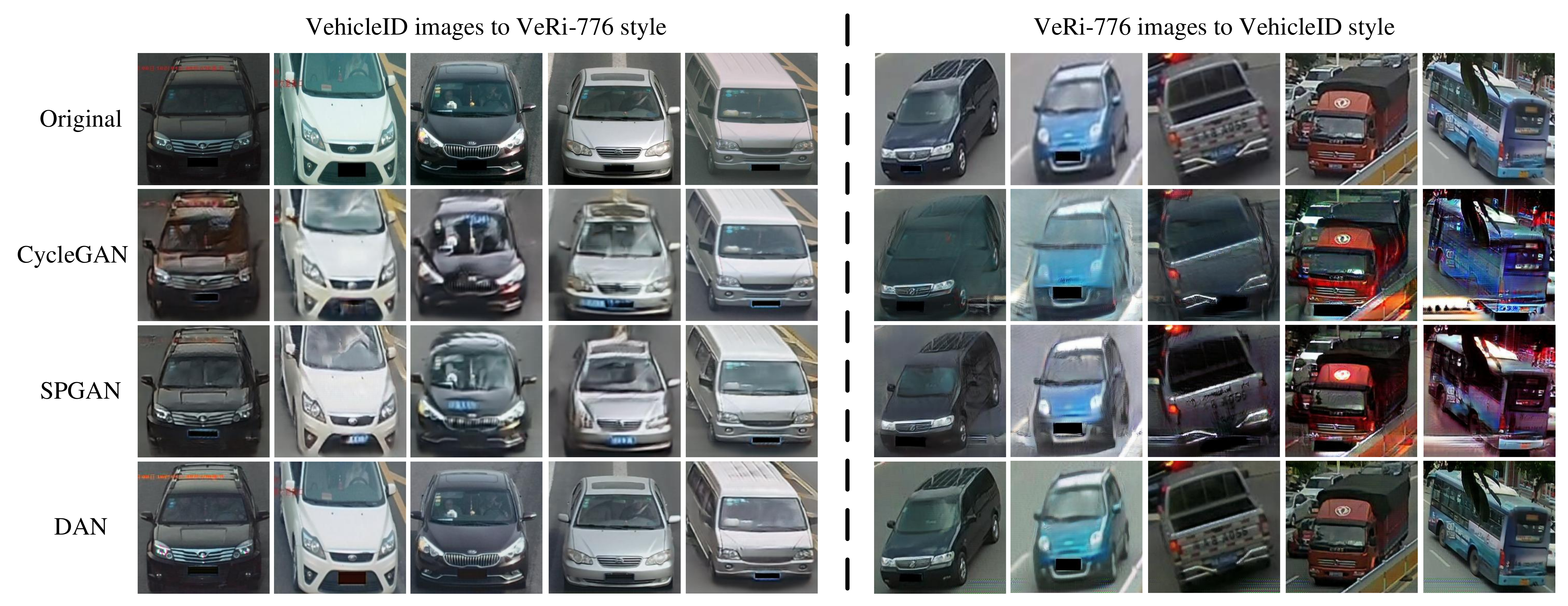}
\caption{The effect of the generated images. The first row is original images. The generated images using CycleGAN, SPGAN lie in the second row and third row respectively. The last row are generated images by DAN.} \label{fig7}
\end{figure}

To demonstrate the effectiveness of our proposed style transferring model, the VehicleID and VeRi-776 are utilized to train the DAN. And CycleGAN and SPGAN are taken as compared methods. Fig.\ref{fig7} is the comparison results. In Fig.\ref{fig7}(a), images are generated with the identity information from the VeRi-776 and the style of VehicleID. In Fig.\ref{fig7}(b), images are generated with the identity information from the VehicleID and the style of VeRi-776. For each group, the first row is the original images in VeRi-776. The second and third rows are generated by CycleGAN and SPGAN, respectively. The last row is generated by the proposed DAN.

From the Fig.\ref{fig7}(a), it could be observed that most images generated by CycleGAN are distorted seriously when transferring images from VehicleID to VeRi-776. And though the SPGAN works better than the CycleGAN, the generated images also have evident deformation. However, employing the proposed DAN, not only is the vehicle color and type information completely preserved, but also the style of the target dataset is learned. As we can see from Fig.\ref{fig7}(a), generated images by DAN have higher resolution and become bright, which learns from the VeRi-776. In the Fig.\ref{fig7}(b), the vehicle information of generated images is better preserved when transferring the VeRi-776 to VehicleID than the VehicleID to VeRi-776. And details in generated images by DAN are also well-retained. What's more, generated images become dark and blurred, which learn from VehicleID by DAN.

\begin{table}[htbp]
\footnotesize
\centering
\setlength{\belowcaptionskip}{10pt}
\caption{Performance of various domain adaptation methods over different reID methods on VeRi-776.}\label{tab1}
\begin{tabular}{p{4cm}|p{1.4cm}|p{1.4cm}|p{1.4cm}}
\hline
 Method &  mAP(\%) & Rank1(\%) & Rank5(\%)\\
\hline
\hline
Supervised Learning &  52.36 & 83.25 & 91.60\\
\hline
Direct Transfer & 19.06 & 55.30 & 67.16\\
CycleGAN+B & 21.45 & 56.37 & 67.16\\
SPGAN+B & 23.27 & 58.05 & 69.31\\
\hline
DAN+B (Ours) & 24.85 & 58.46 & 70.86\\
DAN + ATTNet (DAVR) & 26.35 & 62.21 & 73.66\\
\hline
\end{tabular}
\end{table}

\begin{table}[ht]
\footnotesize
\centering
\setlength{\belowcaptionskip}{10pt}
\caption{Performance of various domain adaptation methods over different reID methods on VehicleID. The mAP (\%) and cumulative matching scores (\%) at rank 1, 5 are listed.}\label{tab2}
\begin{tabular}
{p{4cm}|p{0.8cm}|p{0.8cm}|p{0.8cm}|p{0.8cm}|p{0.8cm}|p{0.8cm}}
\hline
\multirow{2}*{Method} & \multicolumn{3}{c|}{Test size = 800} & \multicolumn{3}{c}{Test size = 1600} \\
\cline{2-7} & mAP & Rank1 & Rank5 & mAP & Rank1 & Rank5\\
\hline
\hline
 Supervised Learning     & 72.40  & 68.04  & 88.03 &70.11 & 66.48 & 84.22\\
 \hline
 Direct Transfer    & 40.05	&35.00	&56.68	&34.90	&30.42	&48.85\\
 CycleGAN+B   & 44.24	&39.39	&60.10	&37.68	&32.97	&53.16\\
 SPGAN+B    &48.27	&42.87&	66.55	&42.51	&37.46	&58.97\\
\hline
 DAN+B (Ours)   & 49.53	&44.44	&66.74	&43.90	&38.97	&59.93\\
 DAN + ATTNet (DAVR)   & 54.01	&49.48	&68.66	&49.72	&45.18	&63.99\\
\hline
\multirow{2}*{Method} & \multicolumn{3}{c|}{Test size = 2400} & \multicolumn{3}{c}{Test size = 3200} \\
\cline{2-7} & mAP & Rank1 & Rank5 & mAP & Rank1 & Rank5 \\
\hline
\hline
 Supervised Learning      &67.97 &64.07	&77.00 &64.30 &61.71 &74.25\\
 \hline
 Direct Transfer    &31.65	&27.28&	44.49 & 29.57 & 25.41 &42.11\\
 CycleGAN+B  &33.17	&28.44	&47.92 & 30.73 & 26.38 &43.84\\
 SPGAN+B    &38.41	&33.54	&53.68 & 35.04 & 30.45 &49.13\\
\hline
 DAN+B (Ours)   &40.07	&35.10	&56.29 & 36.86 & 32.17 & 51.63\\
 DAN + ATTNet (DAVR)  &45.18	&40.71	&59.02 & 42.94 & 38.72 & 55.87\\
\hline
\end{tabular}
\end{table}

\subsubsection{Comparison Methods}
In this paper, we discuss several methods to compare with our proposed method in detail. Supervised learning, which proposed by \cite{ref_article32}, denotes training and testing on target domain, simultaneously. Direct Transfer means directly applying the model trained by images from source domain on the target domain. CycleGAN \cite{ref_article21}, SPGAN \cite{ref_article13} and DAN are employed to translate images from source domain to target domain, and then the generated images are utilized to train reID model. ``B'' represents Baseline \cite{ref_article32} method. ATTNet is our proposed feature learning network.

\subsubsection{Comparison with State-of-the-art Methods}

To verify the effectiveness of the proposed DAN, we conduct several experiments where training sets are images generated from different image translation methods. As shown in Table \ref{tab1} and Table \ref{tab2}, we analyze differences of images translated by CycleGAN, SPGAN and DAN. Compared with CycleGAN, DAN leads to 3.40\% and 2.09\% improvements in mAP and rank-1 on VeRi-776, respectively. On VehicleID, compared with CycleGAN, the gains are 5.05\%, 6\%, 6.66\% and 5.79\% in rank-1 of different test sets, respectively. Though SPGAN has better performance in the stage of image-to-image translation than CycleGAN, it also causes deformation and color distortion in real scenario for vehicle reID task (as Fig.\ref{fig7}). Hence, compared with SPGAN, DAN has 1.34\% and 0.41\% improvements in mAP and rank-1 on VeRi-776. And for different sizes of test sets on VehicleID, DAN has 1.57\%, 1.51\%, 1.56\% and 1.72\% improvements in rank-1, respectively. All of these could demonstrate that the structure of DAN is more stable and could generate suitable samples for training in the target domain. Examples of translated images by DAN are shown in Fig.\ref{fig6}. Besides that, compared with other methods, DAVR could obtain the better performance on both VeRi-776 and VehicleID.

\subsubsection{Comparison Between Supervised Learning and Direct Transfer}
Comparing the supervised learning method with the direct transfer method, it can be clearly observed that a large performance drop when directly utilizing a source-trained model on the target domain. For instance, as shown in Table \ref{tab1}, the baseline model is trained and tested on VeRi-776 achieves 52.36\% in mAP, while dropping to 19.06\% when trained on VehicleID and tested on VeRi-776. From Table \ref{tab2}, it is obvious that a similar drop can be observed when VehicleID is employed as the target domain. When the reID model is trained on VeRi-776 and tested on VehicleID, there are 32.35\%, 35.21\%, 36.32\% and 34.73\% descreases in mAP on different sizes of test sets on VehicleID for the baseline model. The reason behind the performance drop is the bias of data distributions in different domains. This also illustrates that the supervised learning methods trained on source domain can not be utilized on target domain directly.

\begin{table}[htbp]
\footnotesize
\centering
\setlength{\belowcaptionskip}{10pt}
\caption{Comparison of different reID models on VeRi-776.}\label{tab3}
\begin{tabular}{p{4cm}|p{1.4cm}|p{1.4cm}|p{1.4cm}}
\hline
 Methods &  mAP(\%) & Rank1(\%) & Rank5(\%)\\
\hline
\hline
Direct Transfer + B & 19.06 & 55.30 & 67.16\\
CycleGAN + B & 21.45 & 56.37 & 67.16\\
SPGAN + B & 23.27 & 58.05 & 69.31\\
DAN + B & 24.85 & 58.46 & 70.86\\
\hline
Direct Transfer + ATTNet &  23.41 & 59.54 & 70.56\\
CycleGAN + ATTNet & 24.39 & 61.03 & 71.99\\
SPGAN + ATTNet & 25.01 & 61.97 & 71.99\\
DAN + ATTNet (DAVR) & 26.35 & 62.21 & 73.66\\
\hline
\end{tabular}
\end{table}

\begin{figure}[htbp]
\centering
\includegraphics[width=8cm]{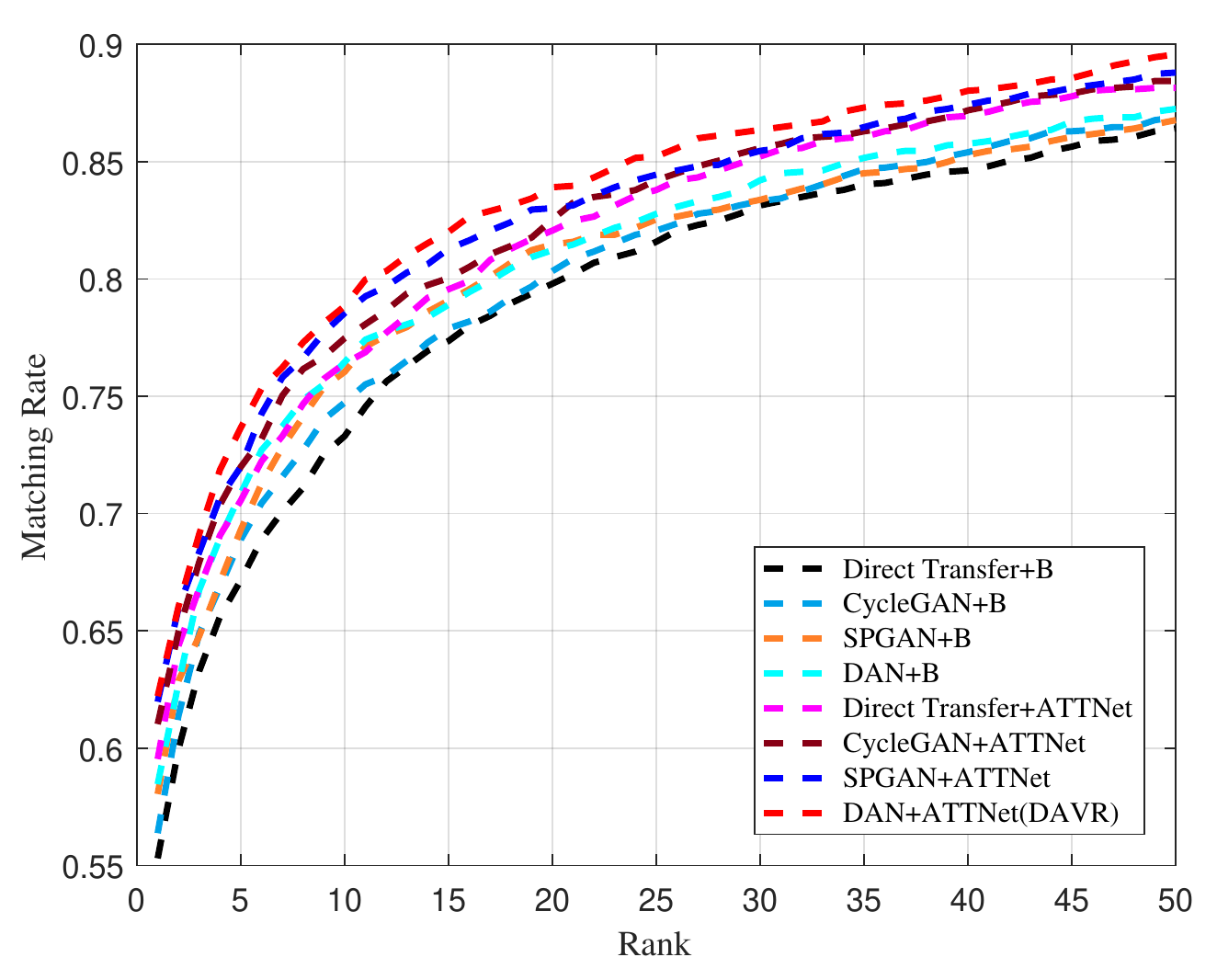}
\caption{ The CMC results of different methods on VeRi-776. } \label{fig8}
\end{figure}

\subsubsection{The Impact of DAN}

Firstly, we utilize DAN to translate labeled images from the source domain to the target domain, and then train the baseline reID model with translated images in a supervised way. As shown in Table \ref{tab1}, when trained on VehicleID and tested on VeRi-776 by baseline method, mAP improves from 19.06\% to 24.85\%. As shown in Table \ref{tab2}, when the reID model is trained on VeRi-776 training set utilizing the baseline method and tested on VehicleID different testing sets, rank-1 accuracy improves from 35\% to 44.44\%, 30.42\% to 38.97\%, 27.28\% to 35.10\% and 25.41\% to 32.17\%, respectively. Through such an image-level domain adaptation method, effective domain adaptation baselines can be learned. This illustrates methods of image-image translation have learned the important style information from the target domain, which could narrow-down the domain gap to a certain degree.

\begin{table}[ht]
\footnotesize
\centering
\setlength{\belowcaptionskip}{10pt}
\caption{Comparison of different reID models on VehicleID. The mAP (\%) and cumulative matching scores (\%) at rank 1, 5 are listed.}\label{tab4}
\begin{tabular}{p{4cm}|p{0.8cm}|p{0.8cm}|p{0.8cm}|p{0.8cm}|p{0.8cm}|p{0.8cm}}
\hline
\multirow{2}*{Methods} & \multicolumn{3}{c|}{Test size = 800(\%)} & \multicolumn{3}{c}{Test size = 1600(\%)} \\
\cline{2-7} & mAP & Rank1 & Rank5 & mAP & Rank1 & Rank5 \\
\hline
\hline
 Direct Transfer + B    & 40.05	&35.00	&56.68	&34.90	&30.42	&48.85\\
 CycleGAN + B   & 44.24	&39.39	&60.10	&37.68	&32.97	&53.16	\\
 SPGAN + B    &48.27	&42.87&	66.55	&42.51	&37.46	&58.97	\\
 DAN + B    & 49.53	&44.44	&66.74	&43.90	&38.97	&59.93	\\
\hline
 Direct Transfer + ATTNet    & 47.97	&43.26	&62.93	&43.94	&39.47	&58.51\\
 CycleGAN + ATTNet    & 46.96	&42.68	&60.72	&43.27	&38.88	&57.44	\\
 SPGAN + ATTNet    & 52.72	&48.25	&67.20	&48.01	&43.44	&63.04\\
 DAN + ATTNet (DAVR)    & 54.01	&49.48	&68.66	&49.72	&45.18	&63.99\\
\hline
\multirow{2}*{Methods} & \multicolumn{3}{c|}{Test size = 2400(\%)} & \multicolumn{3}{c}{Test size = 3200(\%)} \\
\cline{2-7} & mAP & Rank1 & Rank5 & mAP & Rank1 & Rank5\\
\hline
\hline
 Direct Transfer + B    &31.65	&27.28&	44.49 & 29.57 & 25.41 &42.11\\
 CycleGAN + B   &33.17	&28.44	&47.92 & 30.73 & 26.38 &43.84\\
 SPGAN + B    &38.41	&33.54	&53.68 & 35.04 & 30.45 &49.13\\
 DAN + B    &40.07	&35.10	&56.29 & 36.86 & 32.17 & 51.63\\
\hline
 Direct Transfer + ATTNet    &40.42	&35.95	&54.34 & 37.60 & 33.40 & 50.55\\
 CycleGAN + ATTNet    &39.39	&35.09	&53.05 & 37.05 & 33.07 & 49.38\\
 SPGAN + ATTNet    &44.17	&39.51	&59.05 & 41.05 & 36.75 & 54.63\\
 DAN + ATTNet (DAVR)    &45.18	&40.71	&59.02 & 42.94 & 38.72 & 55.87\\
\hline
\end{tabular}
\end{table}

\begin{figure}[htbp]
\centerline{
\subfloat[Test size=800]{\includegraphics[width=2.4in,height=2in]{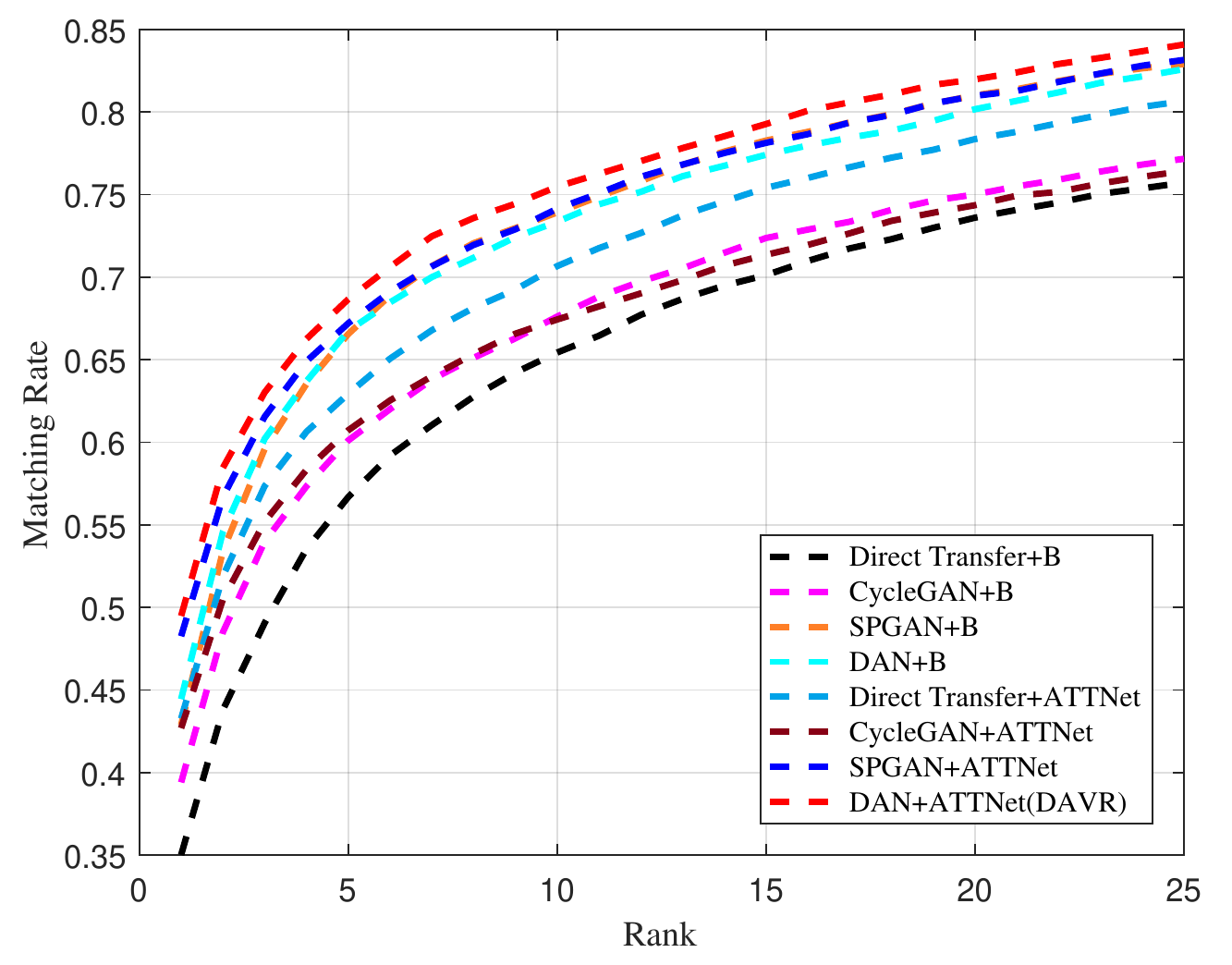}}
\subfloat[Test size=1600]{\includegraphics[width=2.41in,height=2.01in]{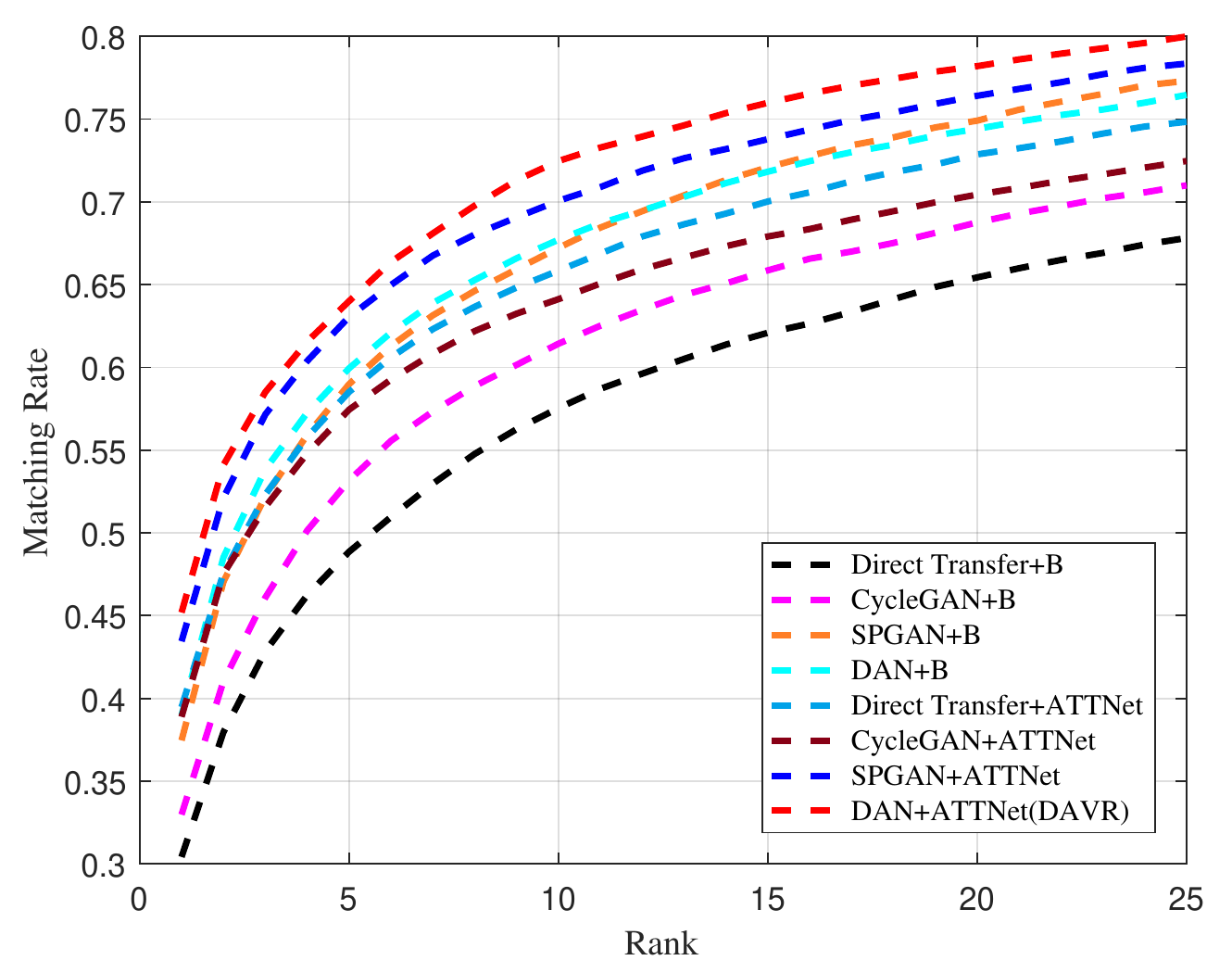}}
}
\centerline{
\subfloat[Test size=2400]{\includegraphics[width=2.4in,height=2.0in]{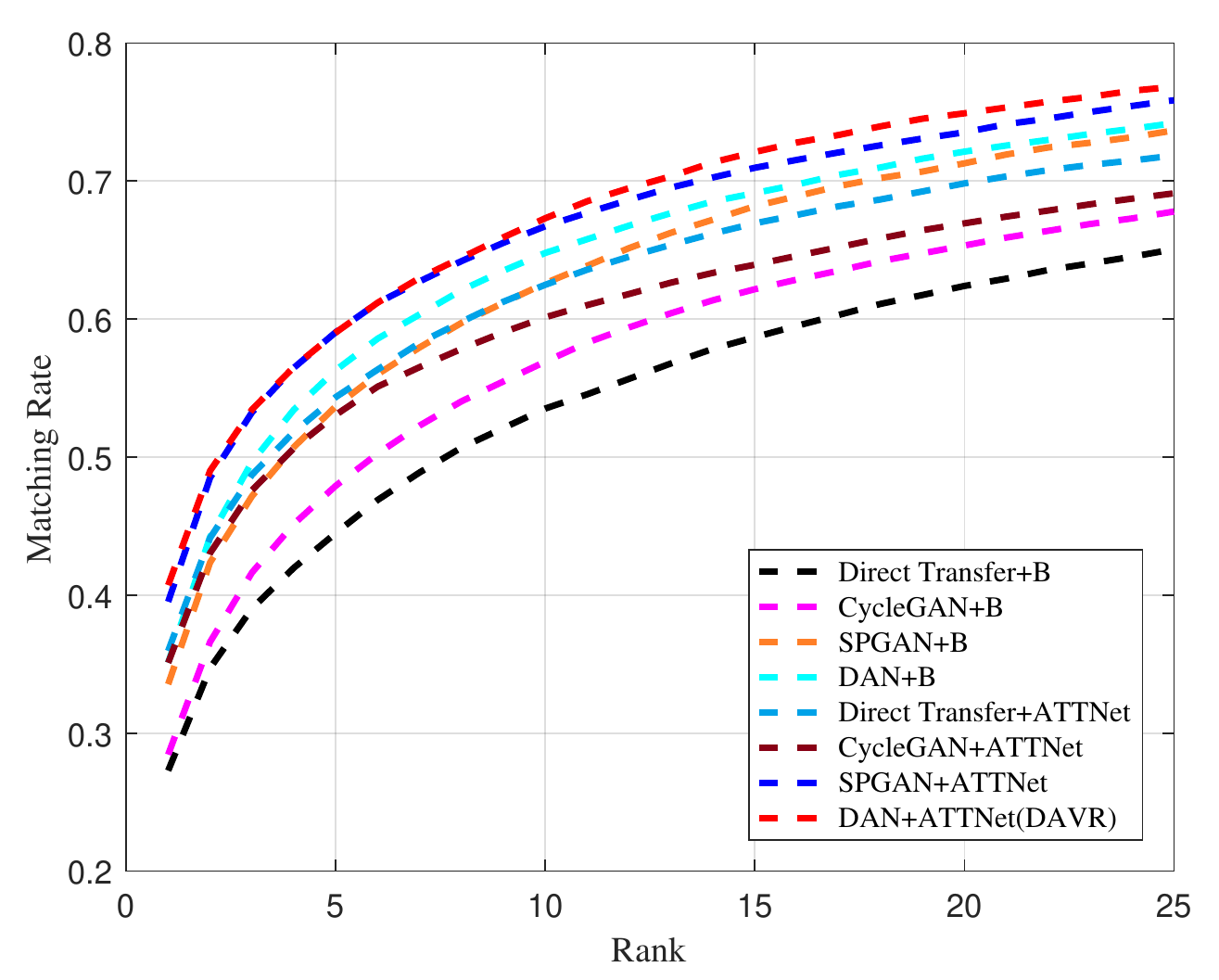}}
\subfloat[Test size=3200]{\includegraphics[width=2.4in,height=2.0in]{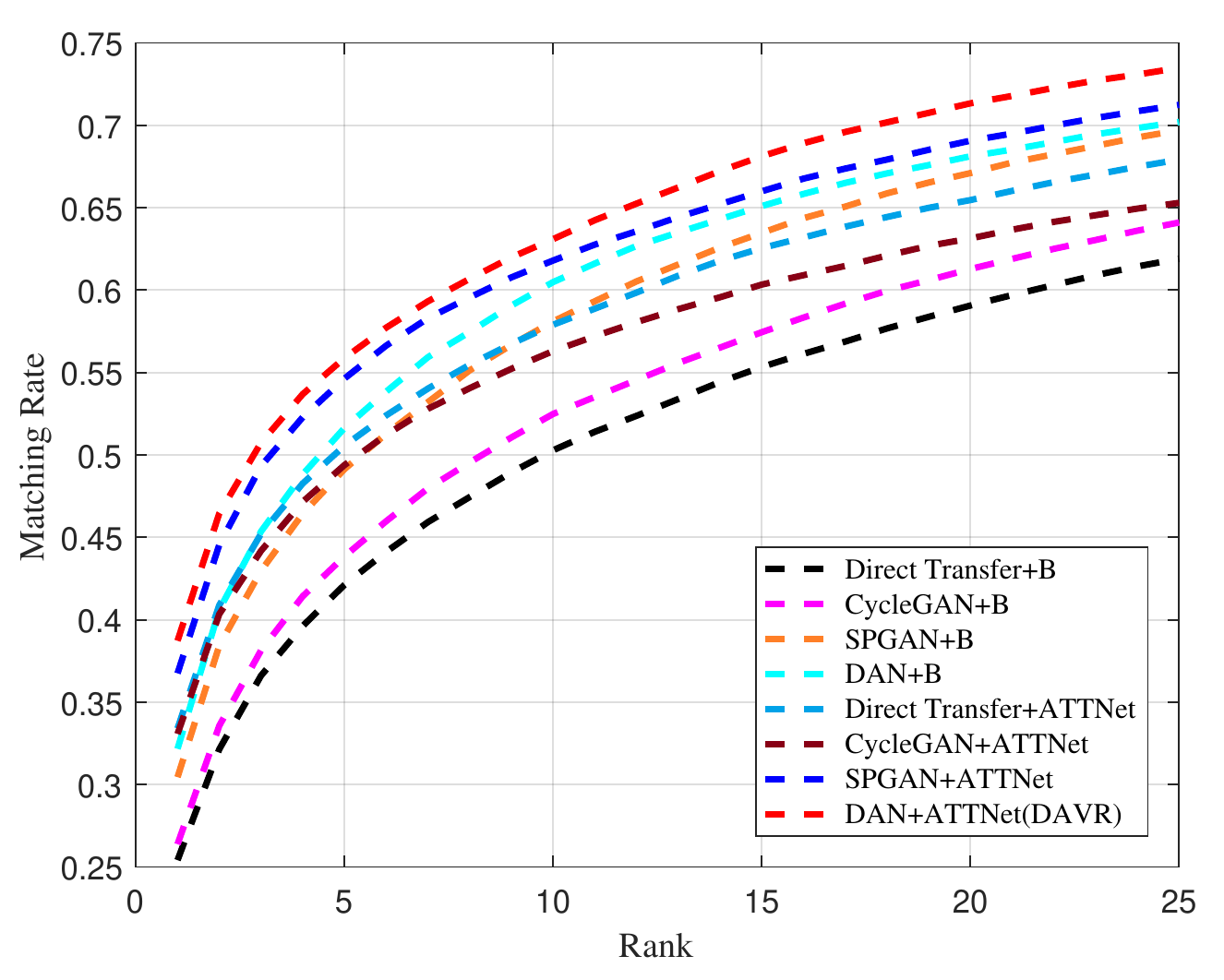}}
}
\caption{The CMC curves of different methods on VehicleID. (a) The results tested on the set with 800 vehicles. (b) The results tested on the set with 1600 vehicles. (c) The results tested on the set with 2400 vehicles. (d) The results tested on the set with 3200 vehicles.} \label{fig9}
\end{figure}

\subsubsection{The Impact of ATTNet}

To further improve re-ID performance on target dataset, we propose ATTNet. Fig.\ref{fig8} and Fig.\ref{fig9} are CMC results on VeRi-776 and VehicleID with different methods. As shown in Fig.\ref{fig8} and Fig.\ref{fig9}, compared to the reID models which are trained by baseline reID model, methods utilizing ATTNet have better performance. For instance, from the Table \ref{tab3} and Table \ref{tab4}, it could be observed that, compared with $Direct\ Transfer+Baseline$, $Direct \ Transfer+ATTNet$ has an 4.35\% increase in mAP when the reID model is trained on VehicleID and tested on VeRi-776. It also has 8.26\%, 9.05\%, 8.67\%, and 7.99\% improvements in rank-1 of different test sets when the model is trained on VeRi-776 and tested on VehicleID. Besides that, it is obvious that compared with the baseline methods, the reID models utilizing the ATTNet have significant improvement for every image translation method. This demonstrates that the reID models which are trained by the proposed ATTNet can better adapt to cross-domain task than the baseline method.

\subsection{Visualization of Results}

\begin{figure}[htbp]
\centerline{
\subfloat[Direct Transfer + Baseline (VeRi-776)]{\includegraphics[width=2.4in,height=2.0in]{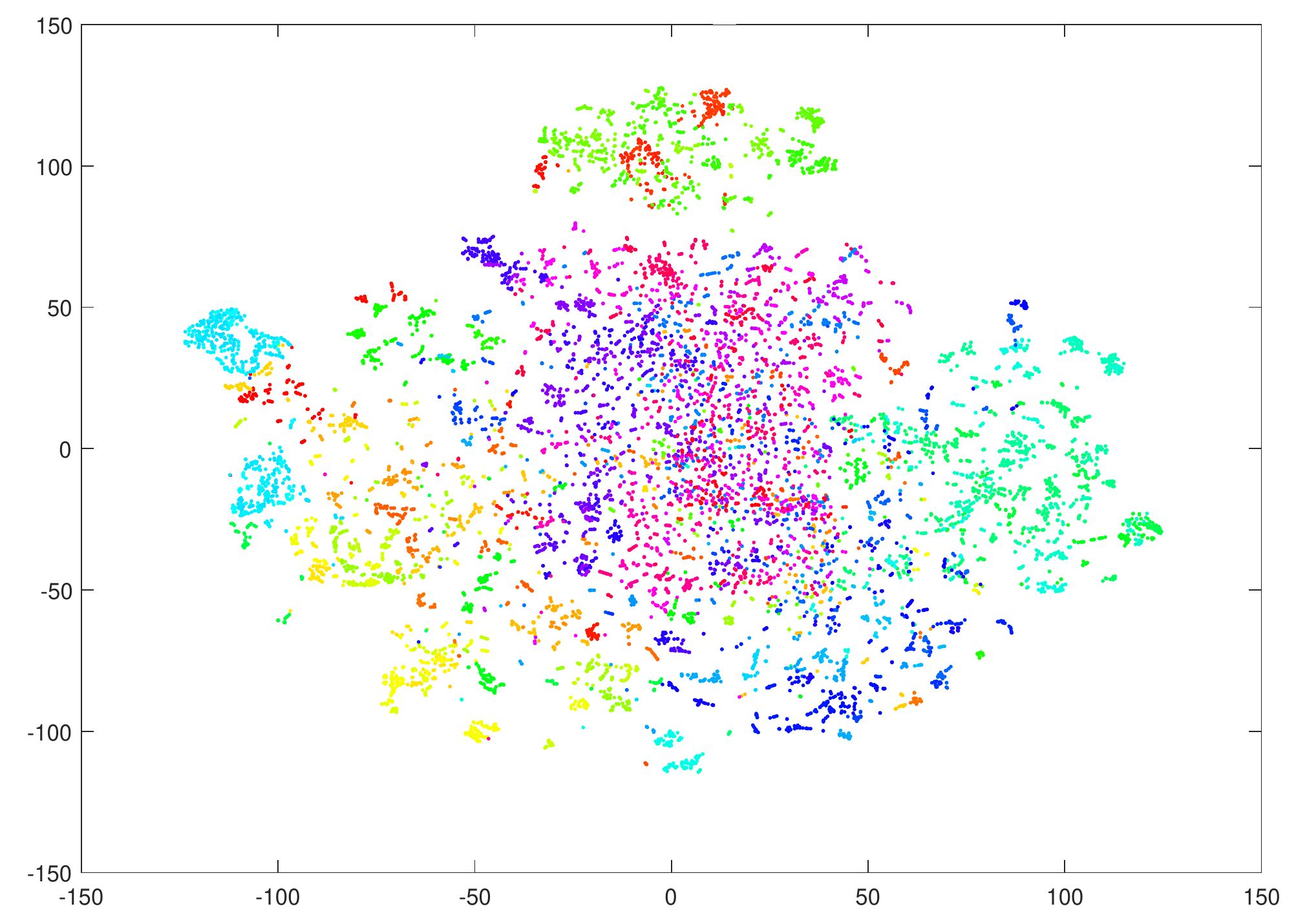}}
\subfloat[DAVR (Ours) (VeRi-776)]{\includegraphics[width=2.4in,height=2.0in]{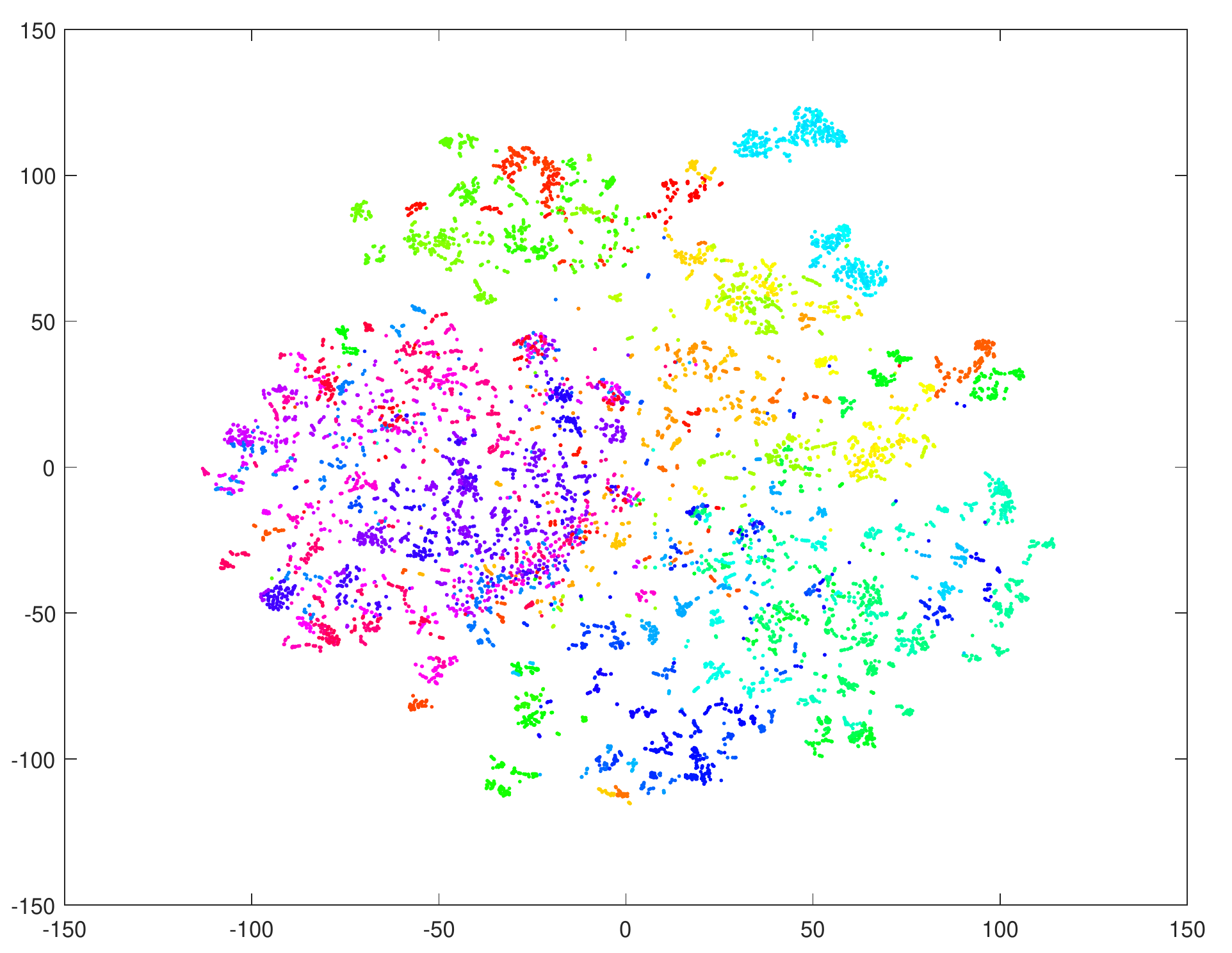}}
}
\centerline{
\subfloat[Direct Transfer + Baseline (VehicleID)]{\includegraphics[width=2.4in,height=2.0in]{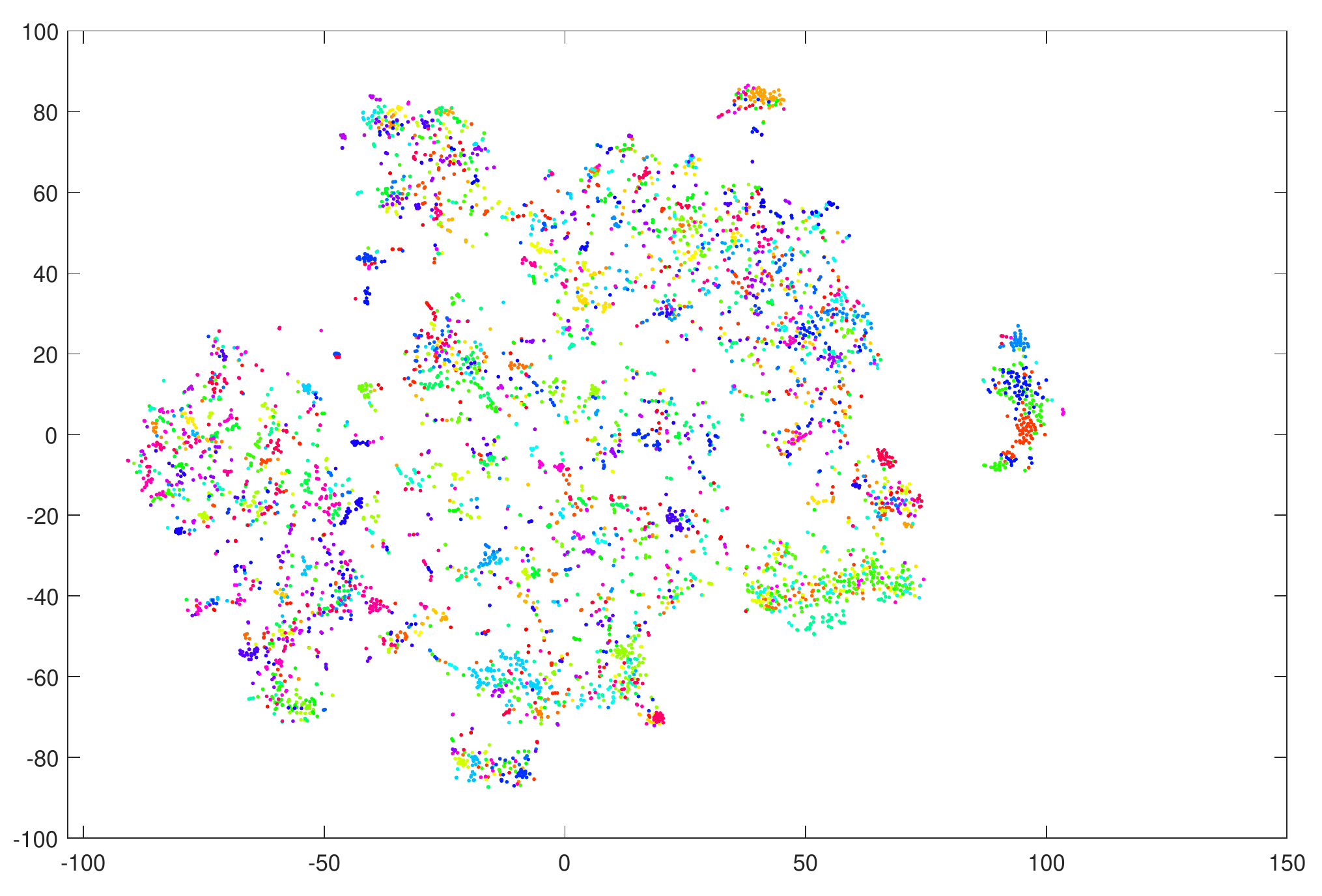}}
\subfloat[DAVR (Ours) (VehicleID)]{\includegraphics[width=2.4in,height=2.0in]{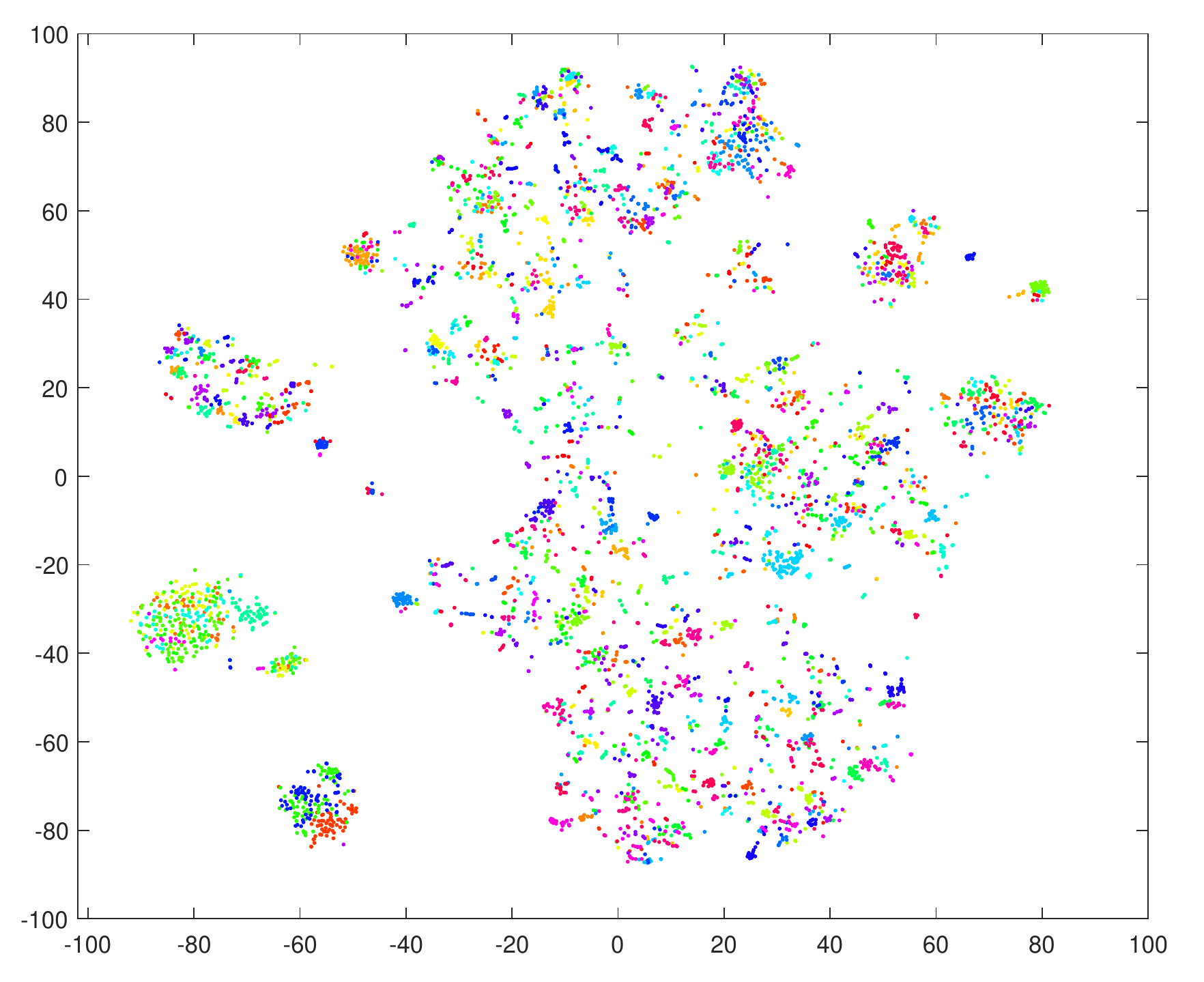}}
}
\caption{Visualization of feature distribution by t-SNE. Different colors represent different vehicle IDs.} \label{fig10}
\end{figure}

To further illustrate the effectiveness of the proposed framework in this paper, some results are visualized. As shown in Fig.\ref{fig10}, we utilize the t-SNE \cite{ref_article33} to visualize the features extracted by different methods. In Fig.\ref{fig10}(a), the reID model is trained on the VehicleID and tested on VeRi-776 with original images by the baseline method \cite{ref_article32}. In Fig.\ref{fig10}(b), the reID model is trained using the ATTNet with images generated by DAN on the VehicleID and tested on VeRi-776. In Fig10(c), the reID model is trained on the VeRi-776 with original images by the baseline method and tested on VehicleID. In Fig.10(d), the reID model is trained using the ATTNet with generated images by DAN on the VeRi-776 and tested on VehicleID. In our experiments, the number of tested vehicle is 200, 800 on VeRi-776 and VehicleID, respectively. From the visualization, we could find that there is a significant improvement compared with the baseline method.

\begin{figure}[htbp]
\centerline{
\subfloat[VeRi-776]{\includegraphics[width=\textwidth]{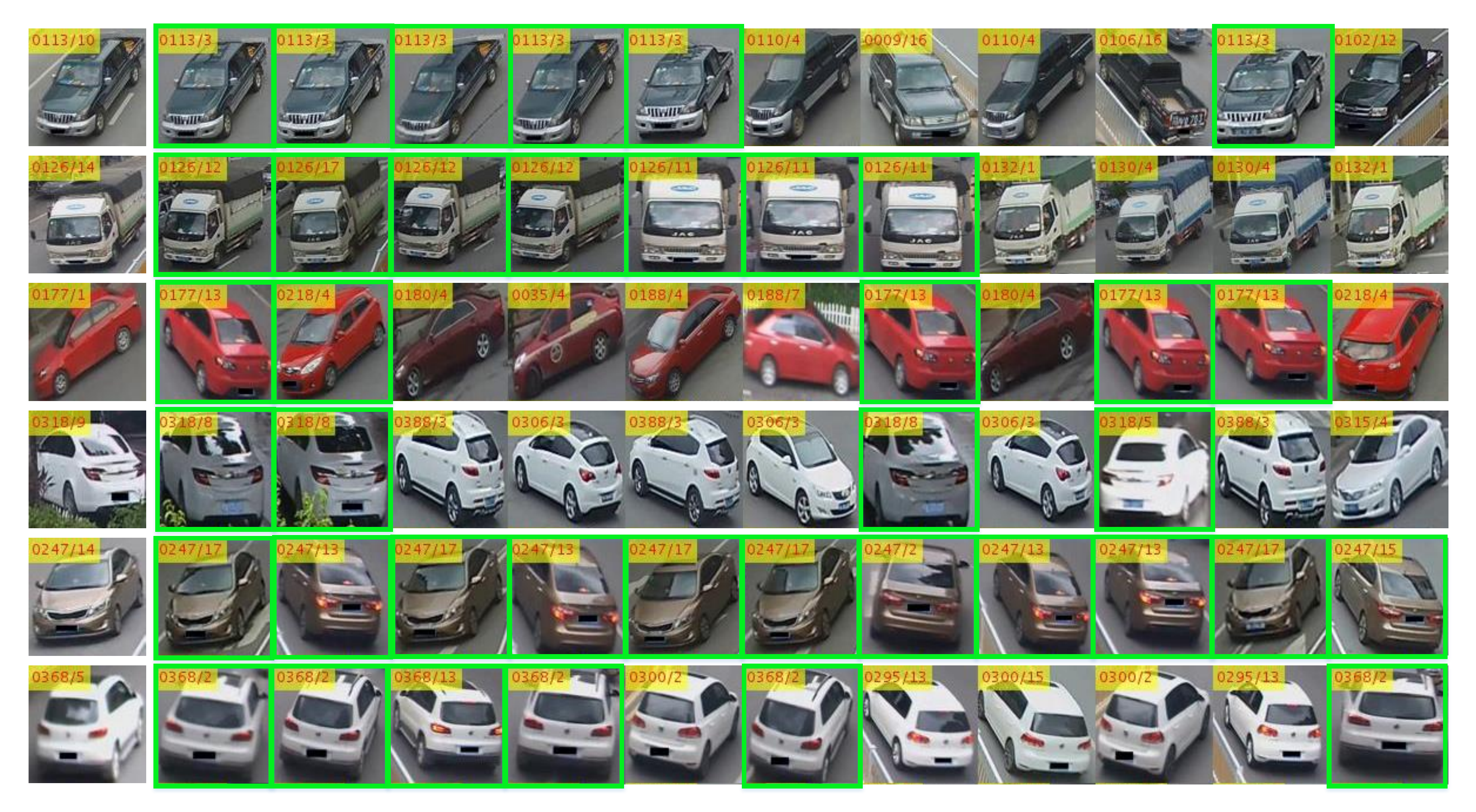}}
}
\centerline{
\subfloat[VehicleID]{\includegraphics[width=\textwidth]{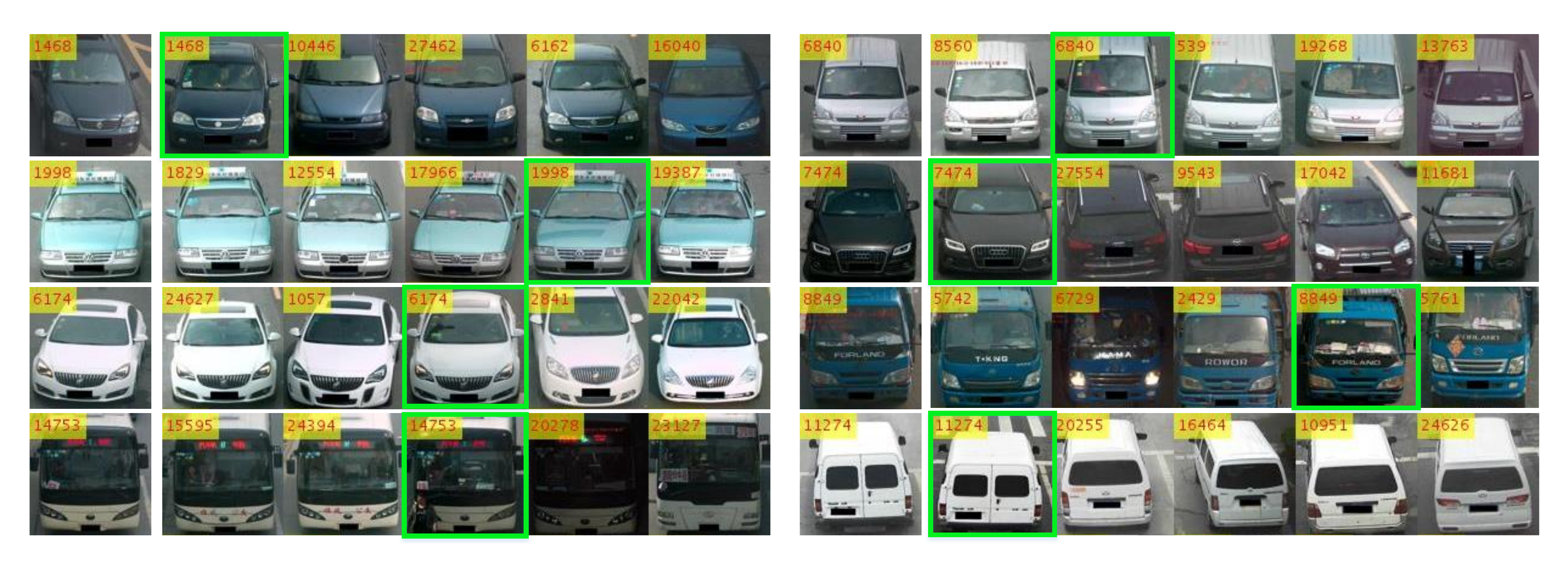}}
}
\caption{The retrieval results on the VehicleID and VeRi-776. (a) The results on VeRi-776. (b) The results on VehicleID.} \label{fig11}
\end{figure}

Examples of vehicle reID results on VeRi-776 and VehicleID by our approach DAVR are shown in Fig.\ref{fig11}. Both in Fig.\ref{fig11}(a) and Fig.\ref{fig11}(b), the left column shows query images while the images of right-hand side are retrieval results obtained by proposed method. The number on the left-top means Vehicle ID/Camera ID for VeRi-776 and Vehicle ID for VehicleID. The same Vehicle ID represents the same vehicle. The Camera ID is the camera number that images are captured. From Fig.\ref{fig11}, we could see that our proposed domain adaptation method in this paper achieves good performance. Specially, in Fig.\ref{fig11}(a), the retrieval results contain different viewpoints and illumination, which could demonstrate that the proposed method has robustness on different conditions.

\section{Conclusion}
In this paper, we propose DAVR, which includes the image-to-image translation and feature learning module for domain adaptation. The DAN is designed to generate the vehicle images, which both preserves the label information of source domain and learns the style of target domain. And then the ATTNet is proposed to train the reID model with generated images. It can be observed from the results that both translation module and feature learning module can achieve good results. And it is obvious that the existing datasets usually only contain several kinds of vehicle images in each camera, which sets a limit for reID task in new domain. Hence, in our future studies, we would aim to utilize the GAN to generate the various viewpoints of vehicle images to expand the dataset and improve the performance of reID model.

\section{Acknowledgements}
This work was supported in part by the National Natural Science Foundation of China Grant 61370142 and Grant 61272368, by the Fundamental Research Funds for the Central Universities Grant 3132016352, by the Fundamental Research of Ministry of Transport of P. R. China Grant 2015329225300, by the Dalian Science and Technology Innovation Fund 2018J12GX037 and Dalian Leading talent Grant, by the Foundation of Liaoning Key Research and Development Program.

%
%

%
%
%

\section{References}

\bibliographystyle{splncs04}
\bibliography{mybibfile}

\end{document}